\documentclass[runningheads]{llncs}

\usepackage{subcaption}
\usepackage{graphicx}
\usepackage{multirow}
\usepackage{xcolor,colortbl}
\usepackage{hyperref}
\hypersetup{
    colorlinks=true,
    linkcolor=blue,
    citecolor=blue,
    urlcolor=blue
}
\definecolor{tmi_blue}{cmyk}{100,0.37,0.0,0.15}

\begin{document}
\title{MedMerge: Merging Models for Effective Transfer Learning to Medical Imaging Tasks}
\titlerunning{MedMerge}
%
%
\author{Ibrahim Almakky, Santosh Sanjeev, Anees Ur Rehman Hashmi, Mohammad Areeb Qazi, Hu Wang, Mohammad Yaqub}  
\authorrunning{Almakky et al.}
\institute{Mohamed bin Zayed University of Artificial Intelligence, Abu Dhabi, UAE \\
    \email{\{firstname.lastname}@mbzuai.ac.ae\}}

%
%
\maketitle              
\begin{abstract}
Transfer learning has become a powerful tool to initialize deep learning models to achieve faster convergence and higher performance. This is especially useful in the medical imaging analysis domain, where data scarcity limits possible performance gains for deep learning models. Some advancements have been made in boosting the transfer learning performance gain by merging models starting from the same initialization. However, in the medical imaging analysis domain, there is an opportunity to merge models starting from different initializations, thus combining the features learned from different tasks. In this work, we propose MedMerge, a method whereby the weights of different models can be merged, and their features can be effectively utilized to boost performance on a new task. With MedMerge, we learn kernel-level weights that can later be used to merge the models into a single model, even when starting from different initializations. Testing on various medical imaging analysis tasks, we show that our merged model can achieve significant performance gains, with up to $7\%$ improvement on the $F_1$ score. The code implementation of this work is available at
\url{www.github.com/BioMedIA-MBZUAI/MedMerge}.


\end{abstract}
\section{Introduction}
A good model initialization plays a critical role in the training and performance of deep learning models. To this end, transfer learning has been established as an effective strategy to boost overall model performance, especially when dealing with small datasets such as the ones available in medical settings \cite{raghu_transfusion_2019}. Despite differences in the nature of tasks and imaging modalities, ImageNet \cite{Krizhevsky2012} has proven to be an effective initialization for many medical imaging tasks \cite{huh_what_2016,matsoukas_what_2022}. The gain in performance is largely due to feature re-use, where the weights learned in the source domain result in features that can readily be used in the target domain \cite{matsoukas_what_2022}. 
Intuitively, transfer learning from a model pre-trained on similar medical imaging tasks that share the same modality and pathology with the downstream task should yield higher performance gains \cite{Xie2018,Ghesu2022Jan}. However, successful transfer learning requires a reasonably large and diverse set of images, prompting efforts to collate large amounts of medical data to effectively pre-train deep learning models \cite{mei_radimagenet_2022}. This is quite challenging in medical imaging, where collating such large amounts of data is often infeasible due to challenges related to data privacy and the availability of expert annotation. 

Model merging has garnered increased attention due to practical possibilities for resource-saving and overcoming data privacy concerns \cite{wortsman_model_2022,Li2023Sep}. Furthermore, with the promise of removing dependence on a single pre-trained model, the merging of model weights can reduce the tendency of a single model to overfit particular samples, thus improving the accuracy, diversity, and robustness of predictions \cite{li_fedbn_2021}. Model weight averaging \cite{Wang2020Apr} is the most direct, simple, and efficient approach to combining several models and has been used effectively in collaborative learning paradigms. However, simple weight averaging can lead to some bias when large differences are present between the pre-training tasks. Model Soups \cite{wortsman_model_2022} implements a greedy algorithm to select fine-tuned models to be merged following a hyperparameter search. The selected models are then aggregated using simple weight averaging, often yielding promising performance gains. FissionFusion \cite{sanjeev2024fissionfusion} adapts Model Soups for medical tasks and overcomes the challenges caused by the rough loss surface nature of models trained on medical datasets. 
However, even if overlooking the computational cost associated with such an extensive search of the hyperparameter space, souping approaches \cite{wortsman_model_2022,sanjeev2024fissionfusion} focus on merging models starting from the same initialization. In medical imaging analysis, it is important to benefit from models that have been pre-trained on different tasks, thus harnessing their capacity for another downstream task. Another approach to model merging, dubbed Fisher Merging, has been proposed to choose parameters that approximately maximize the joint likelihood of the posteriors of models' parameters \cite{matena_merging_2022}. However, they also focus on models starting from the same initialization as Fisher Merging is less performant when applied to models far apart in the parameter space. 

Recent research builds on the intuition, later formalized by \cite{entezari2021role}, that if the permutation invariance of neural networks is taken into account, stochastic gradient descent solutions will likely have no barrier in the linear interpolation between them. This indicates that models permuted to the same loss basin can be merged by averaging their weights. Git Re-basin \cite{ainsworth2022git} introduces three techniques for model permutation to align a second model with a reference model, facilitating their merging in weight space. Their experiments involve models trained on the same data with different initializations. REPAIR \cite{jordan2022repair} enhances Git Re-Basin by incorporating new parameters and adjusting batch norms where applicable. While the permutation concept works for models trained on the same task,
it fails to account for the differences in models trained on disjoint tasks. ZipIt \cite{stoica2023zipit} improves upon the previous works by merging models trained on different datasets into one multi-task model without any additional training.


Focusing on the medical domain and with the aim to harness features learned from different initializations, we propose MedMerge to effectively combine features learned from various tasks. MedMerge can learn kernel-level weights to carry out a weighted average merge of model weights from different initializations, thus learning how to effectively combine features from previous pertaining tasks. The key contributions of this work can be summarized as follows:
\begin{itemize}
    \item We propose a novel method, dubbed MedMerge, to merge models pre-trained on different datasets while effectively utilizing features from different models.
    \item We empirically demonstrate our proposed approach's effectiveness on various medical imaging analysis tasks and show state-of-the-art performance.
    \item We provide insights into the layer-wise feature transfer from source initializations to target tasks and show a correlation between the selection ratio from specific initializations and the performance gain achieved in an ordinary transfer learning setting. 
\end{itemize}




\begin{figure}[t]
    \centering
    \includegraphics[width=\linewidth]{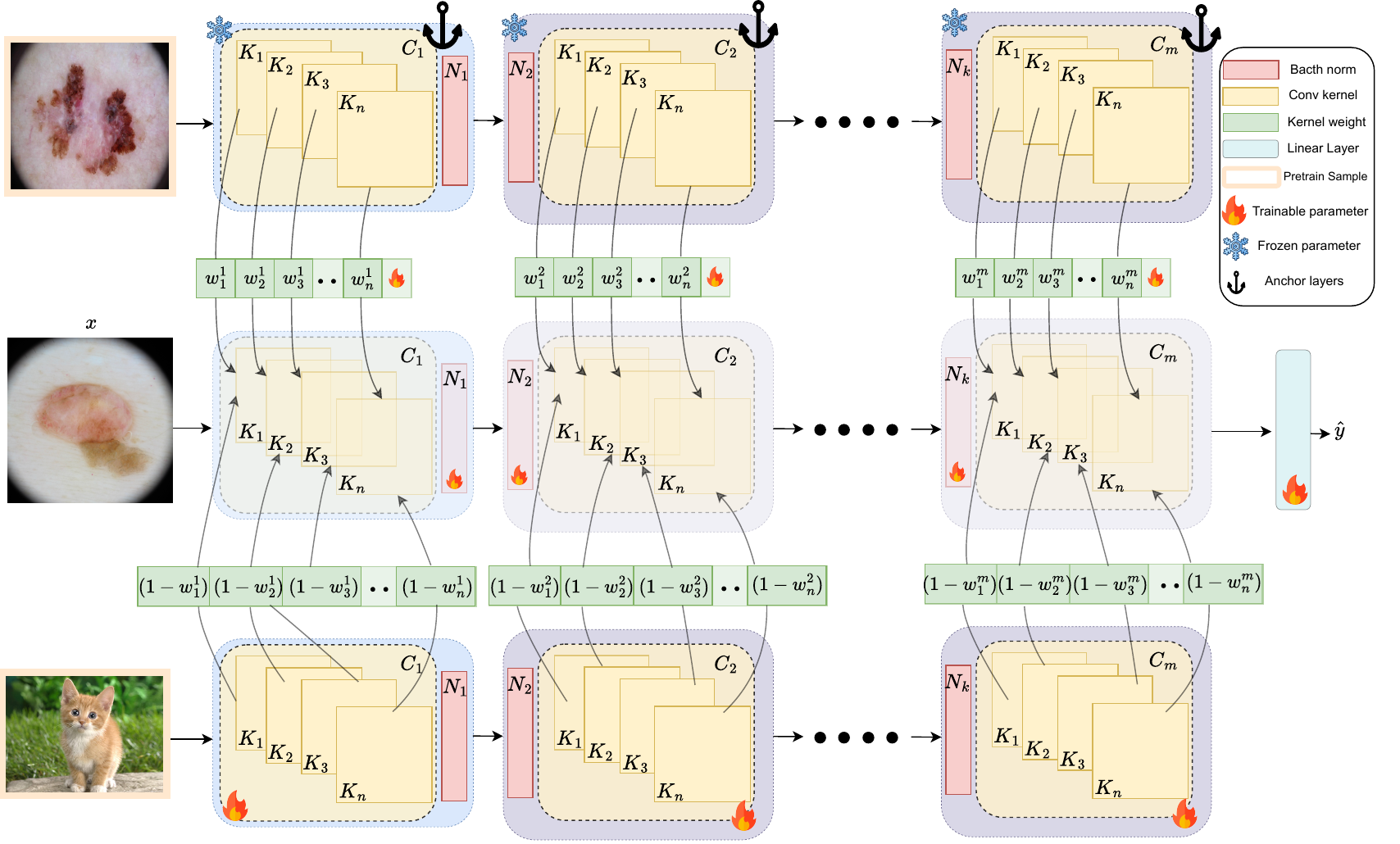}
    \caption{An outline of the proposed MedMege method where the kernels from two pre-trained models are effectively combined towards a new task.}
    \label{fig:methodology}
\end{figure}

\section{Methodology}
\label{sec:method}

\textbf{Problem Formulation.} Let $M$ be a model composed of a feature extraction backbone $B$ and a classification head $H$. The backbone feature extractor $B$ is a Convolutional Neural Network (CNN) with a set of convolutional layers $C \in \{C_1 \dots, C_m\}$. Each layer $C$ has a set of kernels $K = \{K_1, \dots K_n\}$, followed by an activation function. The backbone also has a set of batch normalization layers $\{N_1, \dots, N_k\}$. On the other hand, the classification head $H$ is a fully connected layer that takes the features extracted by $B$ as input and outputs the model predictions. We also consider $\theta^{(i)}_B$ and $\theta^{(i)}_H$ as the backbone and classification head parameters obtained by training model $M$ on dataset $D_i$. This means that the learnt parameters for the convolutional kernel $K_j$ and batch normalization layer $N_j$ can be represented as $\theta_{K_j}^{(i)}$ and $\theta_{N_j}^{(i)}$, respectively. 
\newline

\noindent \textbf{MedMerge.} In this work, we propose MedMerge, a method to effectively combine the weights of two feature extractors $\theta^{(a)}_{B}$ and $\theta^{(b)}_B$, where each has been obtained by pre-training the model architecture $M$ on dataset $D_a$ and $D_b$, respectively. The combined weights will then form $\hat{\theta}^{(q)}$, the effectiveness of which is determined by the performance gains when fine-tuned on unseen dataset $D_q$, which does not intersect with $D_a$ and $D_b$. To accomplish this, we introduce a set of learned weight parameters $\textit{W} = \{w_1, \dots, w_n\}$, where each weight parameter $w_j$ corresponds to a kernel $K_j \in \{K_1, \dots K_n\}$ in the CNN backbone $B$. Those weights are then used to form the convolutional kernels in $\hat{\theta}^{(q)}$, where each kernel is formed as follows: $\hat{\theta}^{(q)}_{K_j} = w_j {\theta}_{K_j}^{(b)} + (1-w_j) {\theta}_{K_j}^{(c)}$, where $\theta^{b}$ and $\theta^{c}$ are the parameters of $M$ after pre-training on datasets $D_b$ and $D_c$, respectively. 

To learn the set of weight parameters $W$, we attach a classification head that takes the joint features from $\hat{\theta}^{(q)}$ at the penultimate layer and outputs the prediction $\hat{y}$ for the new task. \cite{Kumar2022Jan} showed that full fine-tuning distorts the pre-trained features and that linear probing a model before fine-tuning (LP-FT) provides a good initialization for the classification head. Furthermore, due to its limited capability, the linear probe depends heavily on the class separability in the latent feature space, making it useful for evaluating feature generalization \cite{Asano2020Mar}. However, freezing the two pre-trained feature extractors during merging restricts the training process, while unfreezing both models can lead to feature corruption. Therefore, in MedMerge, we propose an anchoring approach where one of the feature extractors is frozen to maintain the grounding of the models during fine-tuning. This anchoring approach is depicted in Fig. \ref{fig:methodology}, and in the ablation study, we empirically show the effect of switching the anchor between a feature extractor trained on medical data versus ImageNet. During the fine-tuning process, $w_j$ is updated based on the gradients for the combined kernels $K_j^{(b)}$ and $K_j^{(c)}$ from both models when calculating the classification loss based on $\hat{y}$.  

Batch normalization \cite{Ioffe2015Feb} plays a critical role when training deep learning models. Specifically for model merging, \cite{li2021fedbn} demonstrated the effectiveness of local batch normalization in alleviating the feature shift between client models in federated learning settings and the impact that has on model aggregation. Thus, in MedMerge, we initialize the batch normalization layers for $\hat{\theta}$ as the mean of the corresponding layers in the aggregated models. Then, unlike frozen convolutional kernels, the batch normalization layers are left unfrozen during the merging stage.

\section{Experimental Setup}

\textbf{Datasets.} We select medical imaging datasets with various modalities and pathologies to adequately assess the efficacy of our approach in comparison with other transfer learning approaches and other state-of-the-art (SOTA) model merging methods. As such, we select two pairs of datasets that share modality or pathology:
\begin{itemize}
    \item[$\circ$] HAM10K \cite{Tschandl2018Aug} and ISIC-2019 \cite{Combalia2019Aug}: Skin lesion datasets with $11,720$ and $25,331$ images, respectively. The HAM10K dataset contains $7$ classes of lesions, while ISIC-2019 contains $8$ classes. We use the official train, validation, and testing split for the HAM10K dataset, while we use a random 60\%, 20\%, and 20\% split for the same sets of the ISIC-2019 dataset. 
    \item[$\circ$] EyePACS \cite{eyePACS} and APTOS \cite{aptos2019}: Diabetic retinopathy datasets with $88,702$ and $3,962$ images, respectively. The two datasets consist of five classes representing the increasing severity of diabetic retinopathy. Since the testing data is not publicly available, we follow a random 80\%, 10\%, and 10\% split for training, validation, and testing, respectively.
\end{itemize}

\noindent \textbf{Implementation Details.} As widely used deep CNN architectures, we primarily use DenseNet-121, ResNet-18, and ResNet-50 as the backbones for all our experiments. We have two training regimes, one for LP and the other during fine-tuning, which are used for MedMerge and LP-FT. For LP, we use random initialization for the fully connected layer of the classification head. We train the model by minimizing the cross entropy loss using the AdamW optimizer \cite{Loshchilov2017Nov} with a learning rate of $10^{-4}$ for $50$ epochs. We also minimize the cross entropy loss during full fine-tuning using the AdamW optimizer, but we decrease the learning rate to $10^{-5}$, and maintain the number of epochs at $50$. All models pre-trained on ImageNet have been specifically trained using ImageNet1K. The models pre-trained on HAM10K, EyePACS, and ISIC-2019 were initialized from ImageNet1K pre-trained parameters. In MedMerge, the kernel-based weights are initialized to a value of $0.5$, which provides an equal merge between the features of the two models.  
All the input images to the models are resized to $224 \times 224$, and to ensure less variability between experiments, we do not conduct any data augmentation strategies during LP or fine-tuning.
All models were trained and tested using NVIDIA A5000 GPUs. The full code, hyperparameters, model weights, and dataset splits will be available upon the paper's acceptance.

\section{Results and Analysis}

\begin{table*}[t]
    \centering 
    \caption{Comparison between the performance achieved by MedMerge, other transfer learning methods, and SOTA merging approaches on the test sets of ISIC-2019, APTOS, and HAM10K. The mean and standard deviation are reported as a result of five random seeds.}
    \resizebox{\textwidth}{!}{
    \begin{tabular}{|c|c|c|c|c|c|c|c|c|c|}
         \hline
         \multirow{2}{*}{Approach} & \multirow{2}{*}{Source} & \multirow{2}{*}{Target} & \multicolumn{2}{c}{DenseNet-121} & \multicolumn{2}{c|}{ResNet-18} & \multicolumn{2}{c|}{ResNet-50}  \\ 
         \cline{4-9}
         &  &  & Acc. & $F_1$ & Acc. & $F_1$ & Acc. & $F_1$ \\
         \hline
         \hline
         Fine-Tune& ImageNet & \multirow{8}{*}{ISIC-2019} & $0.807\pm0.004$ & $0.706\pm0.013$ & $0.776 \pm 0.004$ & $0.648 \pm 0.014$ & $0.772 \pm 0.004$ & $0.632 \pm 0.013$  \\
         Fine-Tune& HAM10K   &  & $0.805 \pm 0.006$ & $0.700 \pm 0.009$ & $0.809 \pm 0.003$ & $0.698 \pm 0.004$ & $0.795 \pm 0.007$ & $0.667 \pm 0.009$ \\
         LP-FT \cite{Kumar2022Jan} & ImageNet &   & $0.806 \pm 0.006$ & $0.701 \pm 0.009$ & $0.777 \pm 0.006$ & $0.637 \pm 0.008$ & $0.772\pm0.015$ & $0.642 \pm 0.023$ \\
         LP-FT \cite{Kumar2022Jan} & HAM10K   &   & $0.810 \pm 0.019$ & $0.709 \pm 0.022$ & $0.797\pm0.011$ & $0.680 \pm0.021$ & $0.789\pm0.006$ & $0.663\pm0.011$ \\
         Average & ImageNet+HAM10K & & $0.818 \pm 0.003$ & $0.724 \pm 0.007$ & $0.803 \pm 0.005$ & $0.682 \pm 0.012$ & $0.805 \pm 0.006$ & $0.657 \pm 0.002$  \\
         Permutation \cite{entezari2021role} & ImageNet+HAM10K   &      & $0.818 \pm 0.002$ & $0.725 \pm 0.006$  & $0.792 \pm 0.004$ & $0.653 \pm 0.006$  & $0.793 \pm 0.005$ & $0.623 \pm 0.024$  \\
         \cellcolor{tmi_blue!10}MedMerge (Ours)  & \cellcolor{tmi_blue!10}ImageNet+HAM10K &  \cellcolor{tmi_blue!10} & \cellcolor{tmi_blue!10} $\textbf{0.837}\pm0.002$  & \cellcolor{tmi_blue!10}$\textbf{0.736}\pm0.006$ & \cellcolor{tmi_blue!10} $\textbf{0.824}\pm0.003$ & \cellcolor{tmi_blue!10} $\textbf{0.706}\pm0.009$  & \cellcolor{tmi_blue!10} $\textbf{0.838}\pm0.002$ & \cellcolor{tmi_blue!10} $\textbf{0.738}\pm0.005$  \\
         \hline
         \hline
         Fine-Tune & ImageNet & \multirow{8}{*}{APTOS} & $0.841\pm0.006$ & $0.702\pm0.009$ & $0.833\pm0.009$ & $0.682\pm0.022$ & $0.840\pm0.009$ & $0.694\pm0.026$  \\
         Fine-Tune & EyePACS  &  & $0.847\pm0.01$ & $0.696\pm0.027$ & $\textbf{0.852}\pm0.009$ & $\textbf{0.710}\pm0.023$ & $0.850\pm0.009$ & $0.700\pm0.020$  \\
         LP-FT \cite{Kumar2022Jan} & ImageNet &  & $0.849\pm0.013$ & $0.698\pm0.035$ & $0.830\pm0.018$ & $0.678\pm0.028$ & $0.829\pm0.026$ & $0.668\pm0.028$   \\
         LP-FT \cite{Kumar2022Jan} & EyePACS  &  & $0.839\pm0.016$ & $0.692\pm0.038$  & $0.833\pm0.021$ & $0.684\pm0.029$ & $0.845\pm0.011$ & $0.697\pm0.026$\\
         Average & ImageNet+EyePACS & & $0.846 \pm 0.010$ & $0.696 \pm 0.023$ & $0.794 \pm 0.010$ & $0.534 \pm 0.034$ & $0.841 \pm 0.004$ & $0.680 \pm 0.015$  \\
         Permutation \cite{entezari2021role} & ImageNet+EyePACS   &      & $0.845 \pm 0.004$ & $0.693 \pm 0.011$ & $0.820 \pm 0.004$ & $0.629 \pm 0.008$ & $0.841 \pm 0.004$ & $0.692 \pm 0.004$ \\
         \cellcolor{tmi_blue!10}MedMerge (Ours) & \cellcolor{tmi_blue!10}ImageNet+EyePACS & \cellcolor{tmi_blue!10} & \cellcolor{tmi_blue!10} $\textbf{0.850}\pm0.009$ & \cellcolor{tmi_blue!10} $\textbf{0.704}\pm0.010$ & \cellcolor{tmi_blue!10} $\underline{0.849}\pm0.006$ & \cellcolor{tmi_blue!10} $\underline{0.689}\pm0.031$  & \cellcolor{tmi_blue!10} $\textbf{0.861}\pm0.007$ & \cellcolor{tmi_blue!10} $\textbf{0.723}\pm0.015$ \\
         \hline
         \hline
         Fine-Tune& ImageNet & \multirow{8}{*}{HAM10K} &  $0.806\pm0.006$ & $0.681\pm0.006$   &  $0.781\pm0.013$ & $0.647\pm0.009$  & $0.784\pm0.019$ & $0.639\pm0.036$ \\
         Fine-Tune& ISIC-2019 & & $0.806\pm0.007$ & $0.697\pm0.022$    & $0.775\pm0.012$ & $0.623\pm0.025$   & $0.791\pm0.006$ & $0.660\pm0.027$  \\
         LP-FT \cite{Kumar2022Jan} & ImageNet &  &  $0.807\pm0.008$ & $0.690\pm0.020$   &  $0.779\pm0.022$ & $0.649\pm0.027$  &  $0.795\pm0.006$ & $0.657\pm0.018$   \\
         LP-FT \cite{Kumar2022Jan} & ISIC-2019 &  &  $0.809\pm0.009$ & $0.701\pm0.016$ &  $0.780\pm0.003$ & $0.637\pm0.005$  &  $0.777\pm0.010$ & $0.640\pm0.016$   \\
         Average & ImageNet+ISIC-2019 & & $0.818 \pm 0.007$ & $0.714 \pm 0.013$ & $\textbf{0.806} \pm 0.003$ & $\textbf{0.695} \pm 0.014$ & $0.813 \pm 0.007$ & $0.699 \pm 0.019$ \\
         Permutation \cite{entezari2021role} & ImageNet+ISIC-2019   &  & $0.816 \pm 0.008$ & $0.703 \pm 0.014$ & $0.788 \pm 0.006$  & $0.648 \pm 0.017$  & $0.795 \pm 0.007$ & $0.641 \pm 0.036$ \\
         \cellcolor{tmi_blue!10}MedMerge (Ours) & \cellcolor{tmi_blue!10}ImageNet+ISIC-2019 & \cellcolor{tmi_blue!10} & \cellcolor{tmi_blue!10}$\textbf{0.820}\pm0.010$ & \cellcolor{tmi_blue!10}$\textbf{0.716}\pm0.020$   & \cellcolor{tmi_blue!10} $\underline{0.802}\pm0.002$ & \cellcolor{tmi_blue!10}$\underline{0.684}\pm0.011$   & \cellcolor{tmi_blue!10}$\textbf{0.813}\pm0.014$ & \cellcolor{tmi_blue!10}$\textbf{0.709}\pm0.026$    \\
         \hline
    \end{tabular}}
    \label{tab:results}
\end{table*}

In Table \ref{tab:results}, we compare the performance of MedMerge with that of Fine-Tuning (FT) and LP-FT \cite{Kumar2022Jan} from source models using a variety of architectures. We also compare with other merging methods, which are simple averaging as well as Permutation \cite{entezari2021role}. MedMerge outperforms other approaches, with significant performance gains on the transfer case from ImageNet and HAM10K to ISIC-2019. We focus on the macro-averaged $F_1$ score as it can better assess the performance of the models with the presence of significant class imbalance, especially in the case of the ISIC-2019 dataset. We can clearly observe the impact of MedMerge's ability to effectively combine the features from the different initializations. In the case of ResNet-18 for APTOS and HAM10K, MedMerge falls slightly below fine-tuning and averaging, respectively. This is likely due to the depth of the ResNet-18 model, which benefits less from the weighted combination of features. 

\begin{figure}[t]
    \centering
    \includegraphics[width=0.95\linewidth]{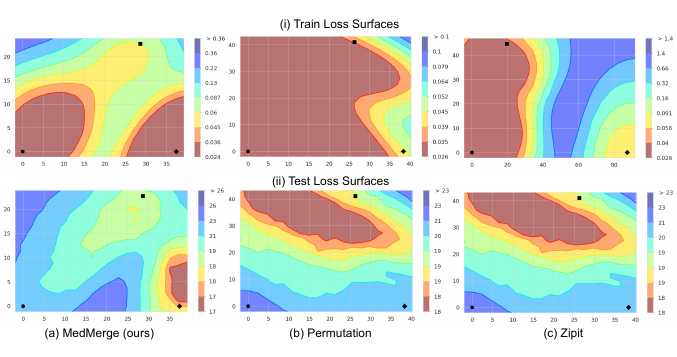}
    \caption{The loss surface plots showing the training and testing error on a two-dimensional slice of
             the error landscapes for MedMerge compared with Permuation \cite{entezari2021role} and Zipit \cite{stoica2023zipit}. This is done for the ResNet-18 model when trained on the ISIC-2019 dataset and starting from the same ImageNet and HAM10K pre-trained models (represented by the square and circle). The diamond represents the merged model.}
    \label{fig:loss_surface}
\end{figure}

MedMerge also outperforms simple weight averaging and Permutation \cite{entezari2021role} approaches, which are affected by the rough loss surface encountered in medical datasets \cite{sanjeev2024fissionfusion}. This is also the case when comparing MedMerge with Zipit \cite{stoica2023zipit}, as shown in Table \ref{tab:ablation}. In Fig. \ref{fig:loss_surface}, we employ \cite{garipov2018loss} to plot the training and testing loss surfaces comparing MedMerge to Permutation \cite{entezari2021role} and Zipit \cite{stoica2023zipit} when combining models for the ISIC dataset. MedMerge is clearly able to not only find a good minimum in the training space but also extend it to the test surface. This highlights the impact of the feature merging on the generalization beyond the training data of the target dataset. On the other hand, Permutation \cite{entezari2021role} and Zipit \cite{stoica2023zipit} get stuck at local minima, which do not generalize well in the test space.

\begin{figure}[t]
    \centering
    \begin{subfigure}[b]{0.49\textwidth}
        \includegraphics[width=\textwidth]{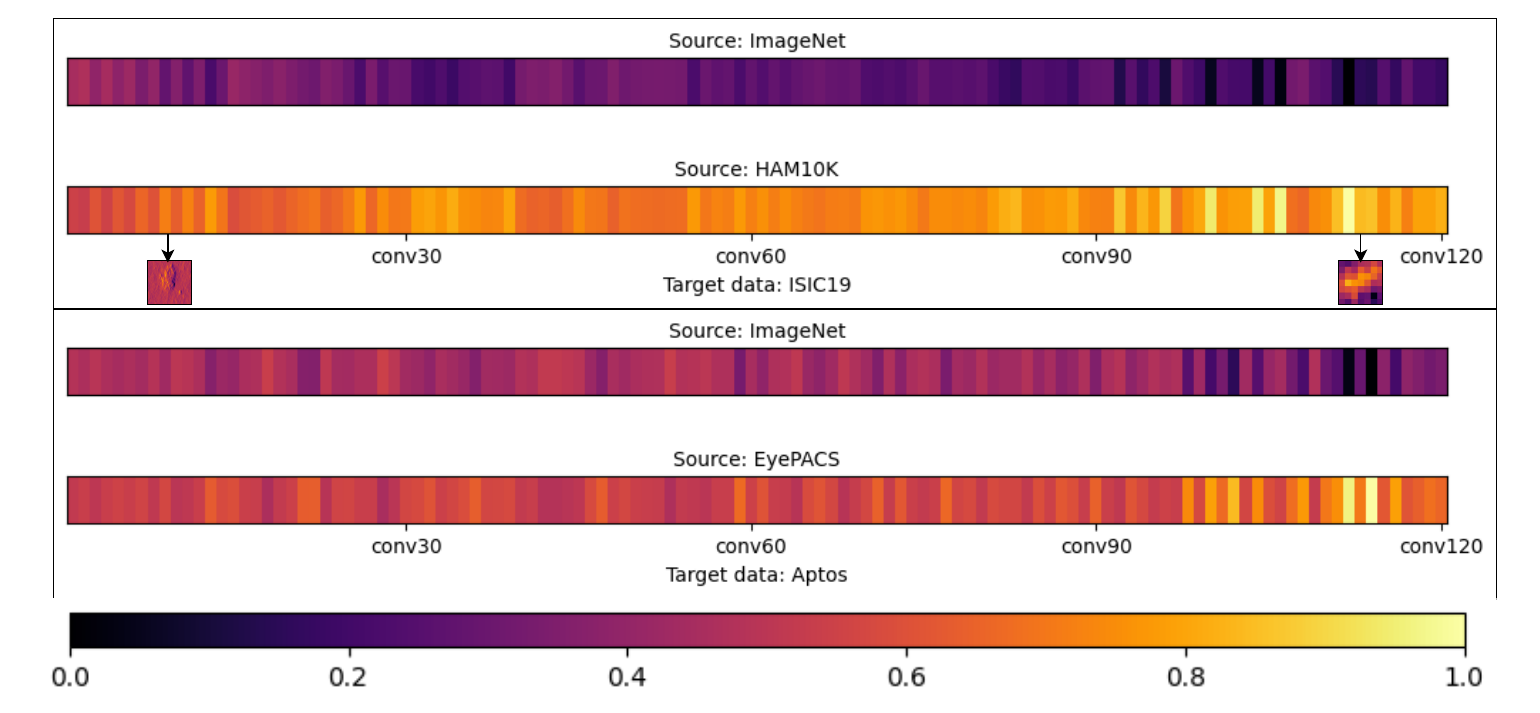}
        \caption{}
        \label{fig:layers_vis_non_zero}
    \end{subfigure}
    \begin{subfigure}[b]{0.49\textwidth}
        \includegraphics[width=\linewidth]{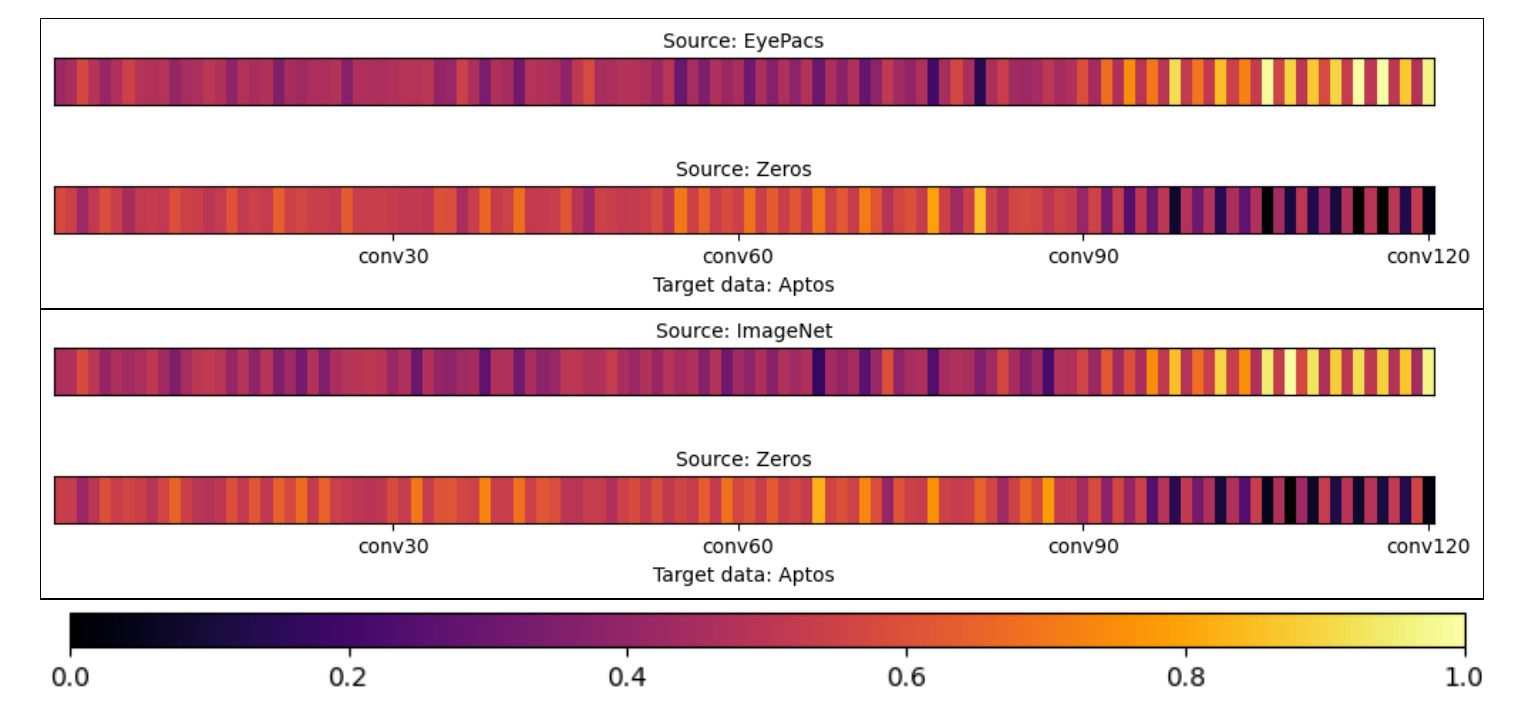}
        \caption{}
        \label{fig:layers_vis_zeros}
    \end{subfigure}
    \caption{Model depth-wise heatmaps showing (a) The mean of learned kernel-based weights at every convolutional layer of the DenseNet-121 backbone. (b) A depth-wise heatmap showing the mean of learned kernel-based weights at every convolutional layer of the DenseNet-121 backbone when merging between one of the source initializations and zeros towards APTOS. }
\end{figure}

To better understand features transferred from source tasks to a target task during our MedMerge merging stage, we freeze the source models and analyze the kernel-based weights learned by merging. We can clearly observe that the overall shift in the weight selection between the two source tasks in Fig. \ref{fig:layers_vis_non_zero} generally corresponds with better FT performance in Table \ref{tab:results}. More specifically, in the ISIC-2019 and EyePACS cases, the kernel-based weights are higher for the datasets sharing the modality with the target, i.e., HAM10K and APTOS, respectively. Moreover, the heatmaps show more weight assigned to layers in the deeper parts of either of the source models toward the classification head. Intuitively, this means that the more abstract features at the end of the model need to belong to a dominant pertaining task. This further highlights the importance of applying different weights at different model layers and shows the effectiveness of the learned MedMerge weights.

\noindent \textbf{Ablation Study.} As summarized in Table \ref{tab:ablation}, we conduct an ablation study surrounding different freezing strategies for MedMerge. We compare fully freezing the feature extractors during merging against unfreezing them and then switching between the anchors. Using the medical feature extractor as an anchor is reliable, even though it does not always yield better performance than other settings. Furthermore, it is evident that unfreezing both feature extractors while merging leads to feature degradation, which leads to a drop in performance.

\begin{table}[t]
    \centering 
    \caption{Comparison between the different different freezing strategies for MedMerge and Zipit \cite{stoica2023zipit} on the ISIC-2019 and APTOS datasets.}
    \resizebox{\textwidth}{!}{
    \begin{tabular}{|l|c|c|c|c|c|c|c|c|}
         \hline
         \multirow{3}{*}{Approach} & \multicolumn{4}{c|}{ISIC-2019}  &  \multicolumn{4}{c|}{APTOS} \\
         \cline{2-9}
         & \multicolumn{2}{c|}{ResNet-18} &  \multicolumn{2}{c|}{ResNet-50} & \multicolumn{2}{c|}{ResNet-18} &  \multicolumn{2}{c|}{ResNet-50} \\
         \cline{2-9}
         & Acc & $F_1$ & Acc & $F_1$   & Acc & $F_1$  & Acc & $F_1$  \\
         \hline
         \hline
          Zipit \cite{stoica2023zipit} & $0.786 \pm 0.004$ & $0.634 \pm 0.011$ & $0.783 \pm 0.006$ & $0.594 \pm 0.035$ & $0.802 \pm 0.014$ & $0.568 \pm 0.040$ & $0.834 \pm 0.009$ & $0.685 \pm 0.014$ \\
         \hline
         \hline
         MedMerge - Fully Unfrozen & $0.809\pm0.002$ & $0.690\pm0.009$  & $0.829\pm0.007$ & $0.724\pm0.017$  & $0.843\pm0.006$ & $0.684\pm0.02$  & $0.838\pm0.013$ & $0.666\pm0.017$   \\
         MedMerge - Fully Frozen & $0.799\pm0.004$ & $0.667\pm0.009$  & $0.812\pm0.004$ & $0.686\pm0.009$    & $0.838\pm0.008$ & $0.683\pm0.018$ & $\textbf{0.868}\pm0.004$ & $\textbf{0.736}\pm0.008$  \\
         MedMerge - ImageNet Anchor & $\textbf{0.827}\pm0.004$ & $\textbf{0.710}\pm0.002$ & $0.832\pm0.006$ & $0.727\pm0.015$  & $0.845\pm0.012$ & $0.680\pm0.014$  & $0.861\pm0.008$ & $0.727\pm0.016$  \\
         MedMerge - Medical Anchor & $0.824\pm0.003$ & $0.706\pm0.009$  & $\textbf{0.838}\pm0.002$ & $\textbf{0.738}\pm0.005$  & $\textbf{0.849}\pm0.006$ & $\textbf{0.689}\pm0.031$ & $0.859\pm0.008$ & $0.718\pm0.017$  \\

         \hline
    \end{tabular}}
    \label{tab:ablation}
\end{table}

It is known that zero initialization will not contribute to extracting meaningful features from the target dataset, which implies that learning to pick zeros over a feature is likely because that feature is detrimental. In Fig. \ref{fig:layers_vis_zeros}, we analyze the learned weights when combining a source initialization with zeros as an initialization. In such a way, we can see a pattern of deeper layers where the kernel-level weights are alternating towards the non-zero initialization. The alternating pattern can be attributed to the nature of the blocks in the DenseNet-121 backbone. This further supports the argument regarding the importance of different merge weights between layers. 
We also tested with freezing batch normalization layers and zero initialization during the merging stage. This led to a drop in the model's performance during both the merging and FT stages. This implies that the mean of the batch normalization layers from the two source initializations is less effective for the target task.

\noindent \textbf{Computational Cost.} The computational cost associated with a parameter-level weighted averaging approach can be high, mainly because of the additional weight parameters required and the training process that entails computing gradients on the new target data. Therefore, in MedMerge, we opt for kernel-level learned weights rather than parameter level, making it more efficient. Furthermore, learning the kernel-level weights is only required during the merging stage, after which we can use the learned weights to combine the models before moving on to the fine-tuning process.

\section{Conclusions and Future Work}
We address an important opportunity in the medical imaging analysis domain: to use features learned from models pre-trained on different tasks to help boost performance on a target task. To accomplish this, we propose MedMerge to merge models starting from different initializations effectively. In MedMerge, we show that learning kernel-level weights to combine models from different initializations results in a merged model that can outperform the performance achieved by fine-tuning, linear probing then fine-tuning (LP-FT), and state-of-the-art merging methods. We demonstrate the performance gains of MedMerge on various medical imaging analysis tasks and conduct layer-wise analysis of the merging process. We conclude that this work mainly focuses on CNN backbones, but in the future, we intend to investigate this aggregation approach in Transformer-based architecture. Furthermore, it would be important to investigate approaches to combining more than two model initializations simultaneously.

%
%
%
\bibliographystyle{splncs04}
\bibliography{references}
\end{document}


\title{MedMerge: Merging Models for Effective Transfer Learning to Medical Imaging Tasks}
\titlerunning{MedMerge}

\maketitle  

\begin{table}
    \centering 
    \caption{Comparison between the performance achieved using the ResNet-50 model by our learnt kernel-based averaging method and other transfer learning methods on the test sets of ISIC-2019, APTOS, and RSNA-Pneumonia datasets. MedMerge achieves $F_1$ performance gains on the ISIC-2019 and APTOS datasets, while falling very slightly short on RSNA-pneumonia.}
    \resizebox{\textwidth}{!}{
    \begin{tabular}{|c|c|c|c|c|c|}
         \hline
         Approach  & Source & Target & Acc. & $F_1$ \\
         \hline
         \hline
         FT     & ImageNet & ISIC-2019 & $0.771$ & $0.634$ \\
         \rowcolor[gray]{0.95} FT     & HAM10K   &  ISIC-2019    & $0.803$  & $0.664$ \\
         LP-FT  & ImageNet & ISIC-2019                           & $0.776$ & $0.637$ \\
         \rowcolor[gray]{0.95} LP-FT  & HAM10K   & ISIC-2019     & $0.781$ & $0.642$ \\
         \rowcolor[gray]{0.9} MedMerge (Ours)  & ImageNet+HAM10K & ISIC-2019 & $\textbf{0.821}$ & $\textbf{0.709}$ \\
         \hline
         \hline
         FT  & ImageNet & APTOS & $0.837$ & $0.674$ \\
         \rowcolor[gray]{0.95} FT  & EyePACS  & APTOS & $\textbf{0.853}$ & $0.702$  \\
         
         LP-FT  & ImageNet & APTOS & $0.837$ & $0.685$ \\
         \rowcolor[gray]{0.95} LP-FT  & EyePACS  & APTOS & $0.852$ & $0.706$ \\
         \rowcolor[gray]{0.9} MedMerge  (Ours) & ImageNet+EyePACS & APTOS & $0.850$ & $\textbf{0.710}$ \\
         \hline
         \hline
         FT  & ImageNet &  RSNA-Pneumonia & $0.946$ & $0.944$ \\
         \rowcolor[gray]{0.95} FT  & CheXpert & RSNA-Pneumonia & $0.946$ & $0.942$ \\
         
         LP-FT  & ImageNet & RSNA-Pneumonia & $0.936$ & $0.933$ \\
         \rowcolor[gray]{0.95} LP-FT  & CheXpert & RSNA-Pneumonia & $\textbf{0.951}$ & $\textbf{0.948}$ \\
         \rowcolor[gray]{0.9} MedMerge  (Ours)  & ImageNet+CheXpert & RSNA-Pneumonia & $0.945$ & $0.942$ \\
         \hline
    \end{tabular}}
    \label{tab:results}
\end{table}